\pdfoutput=1
\documentclass[11pt]{article}

\usepackage{EMNLP2022}
\usepackage{times}
\usepackage{latexsym}
\usepackage{multirow}
\usepackage{tabularx}
\usepackage{amsmath}
\usepackage{amssymb}
\usepackage{bm}
\usepackage{subfigure}
\usepackage{latexsym}
\usepackage{arydshln}
\usepackage{microtype}
\usepackage[switch]{lineno}
\usepackage{helvet} % DO NOT CHANGE THIS
\usepackage{courier}  % DO NOT CHANGE THIS
\usepackage{graphicx} % DO NOT CHANGE THIS
\usepackage{natbib}  % DO NOT CHANGE THIS AND DO NOT ADD ANY OPTIONS TO IT
\usepackage{caption} % DO NOT CHANGE THIS AND DO NOT ADD ANY OPTIONS TO IT

\usepackage{times}
\usepackage{latexsym}

\usepackage[T1]{fontenc}
\usepackage[utf8]{inputenc}

\usepackage{microtype}

\title{Improving Zero-Shot Multilingual Translation with \\ Universal Representations and Cross-Mappings
}

\author{Shuhao Gu\textsuperscript{\rm 1,2}, Yang Feng\textsuperscript{\rm 1,2}\thanks{$^*$Corresponding author: Yang Feng. \newline \indent Reproducible code: https://github.com/ictnlp/Zero-MNMT. } \\ 
\textsuperscript{\rm 1} Key Laboratory of Intelligent Information Processing,\\ Institute of Computing Technology, Chinese Academy of Sciences (ICT/CAS)\\
\textsuperscript{\rm 2} University of Chinese Academy of Sciences\\
{ \{gushuhao19b,fengyang\}@ict.ac.cn}}

\begin{document}
\maketitle
\begin{abstract}
The many-to-many multilingual neural machine translation can translate between language pairs unseen during training, i.e., zero-shot translation. 
Improving zero-shot translation requires the model to learn universal representations and cross-mapping relationships to transfer the knowledge learned on the supervised directions to the zero-shot directions. 
In this work, we propose the state mover's distance based on the optimal theory to model the difference of the representations output by the encoder.
Then, we bridge the gap between the semantic-equivalent representations of different languages at the token level by minimizing the proposed distance to learn universal representations. 
Besides, we propose an agreement-based training scheme, which can help the model make consistent predictions based on the semantic-equivalent sentences to learn universal cross-mapping relationships for all translation directions. 
The experimental results on diverse multilingual datasets show that our method can improve consistently compared with the baseline system and other contrast methods. The analysis proves that our method can better align the semantic space and improve the prediction consistency.%\footnote{We will release our codes and data resources after anonymity period.} 
%The analysis prove that our method can actually better match the semantic space of different languages. 
%faces many challenges, one of which is how to further align the semantic space so the encoder can produce equivalent state sequences for the synonymous sentences. To address this problem, we propose state mover's distance based on the optimal transport theory to model the difference between two state sequences and bridge the representation gaps by minimizing the distance. 
%to use the optimal transportation theory to model the distance between two state sequences 
%and we match the source and target sentence by reducing the corresponding distance. 
%Besides, we also propose an agreement-based training method to let the decoder make consistent predictions based on the equivalent state sequences.
%representations of the source and target sentence. %We conducted experiments on serval multilingual datasets with. q
%We can bridge the gap between the semantic representations of the synonymous sentences at the token level by minimizing the proposed distance. 
\end{abstract}

\section{Introduction}
The many-to-many multilingual neural machine translation (NMT)~\cite{HaNW16,FiratCB16,JohnsonSLKWCTVW17,GuHDL18,abs-2010-11125,ZhangWTS20} model can support multiple translation directions in a single model. The shared encoder encodes the input sentence to the semantic space, and then the shared decoder decodes from the space to generate the translation of the target language. This paradigm allows the model to translate between language pairs unseen during training, i.e., zero-shot translation.
Zero-shot translation can improve the inference efficiency and make the model require less bilingual training data. 
%Despite the potential advantages, improving zero-translation faces many challenges.  
%the zero-shot directions often suffer poor translation quality.
%because of the lack of direct supervision signals.
Performing zero-shot translation requires universal representations to encode the language-agnostic features and cross-mapping relationships that can map the semantic-equivalent sentences of different languages to the particular space of the target language. 
In this way, the model can transfer the knowledge learned in the supervised translation directions to the zero-shot translation directions.
%the encoder to convert the input sentence into a language-agnostic semantic representation, and then the decoder maps the semantic representation to the target language space to generate the translation. 
%Only in this way can the knowledge learned in the supervised translation directions be transferred to the zero-shot translation directions. 
However, the existing model structure and training scheme cannot ensure the universal representations and cross-mappings because of lacking explicit constraints.
%model to learn a unified semantic space and mapping relationship for each translation direction. 
Specifically, the encoder may map different languages to different semantic subspaces, and the decoder may learn different mapping relationships for different source languages, especially when the model possesses high capacity. 
%be map the semantic representations of different languages to different subspaces of the target language, especially when the model possesses high capacity. 
%Consequently, only limited knowledge learned in the supervised translation directions can be transferred to the zero-shot translation directions, which leads to poor zero-shot translation results eventually.
%This makes the translation quality of zero-shot often poor.

Many researchers have made their attempts to solve this problem. \citet{PhamNHW19} propose to compress the output of the encoder into a consistent number of states to only encode the language-independent features. \citet{abs-1903-07091} add a regularizing loss to maximize the similarities between the sentence representations of the source and target sentences. \citet{PanWWL20} propose contrastive learning schemes to minimize the sentence representation gap of similar sentences and maximize that of irrelevant sentences. 
All the above work tries to minimize the representation discrepancies of different languages at the sentence level, bringing two problems for NMT.
%to align the semantic space.
%However, they all convert the state sequences output by the encoder to a length-fixed representation to make the sentences from different languages comparable, which brings two problems for NMT. 
Firstly, these work usually get the sentence-level representation of the encoder output by max-pooling or averaging, which may potentially ignore the sentence length, word alignment relationship, and other token-level information. Secondly, regularizing sentence representation mismatches to the working paradigm of the NMT model, because the decoder directly performs cross attention on the whole state sequences rather than the sentence representation. Besides, all the above work focuses on the encoder side and cannot help learn the universal mapping relationship for the decoder. 
%Therefore, although the averaged sentence-level representation has been proved effective for many tasks~\cite{ReimersG19}, simply bridging the gap between the averaged representations is not enough for the encoder to produce equivalent state sequences.
%bridging the gap between the averaged sentence-level representations  during decoding. 
%simply bridging the gap between the length-fixed representations does not necessarily make the sequence representations equivalent to the decoder.

Given the above, we propose a method to learn the universal representations and cross-mappings to improve the zero-shot translation performance. 
%directly bridging the gap between the state sequences of the source and target sentences to further align the semantic space.
%and help the decoder make consistent predictions based on them. 
%To achieve this, we first feed both the source and target sentences to the encoder to get the corresponding state sequences. 
Based on the optimal transport theory, we propose state mover's distance (SMD) to model the differences of two state sequences at the token level.
%which is measured as the minimum cost that each state vector of one state sequence need to "travel" to reach a certain state vector of another state sequence, to model the differences of two state sequences.
%Then we use the optimal transport theory to model their difference, which is measured as the minimum amount distance that each state vector of one sequence need to "travel" to reach a certain state vector of another sequence. 
To map the semantic-equivalent sentences from different languages to the same place of the semantic space, we add an auxiliary loss to minimize the SMD of the source and target sentences. 
%Therefore, the encoder can produce semantic-equivalent representations for the synonyms sentences.
%term to the objective function to assist the encoder in producing equivalent state sequences for the semantic-equivalent sentences from different languages.  
%help regularize the encoder output representation, so the encoder can produce equivalent state sequences for synonymous sentences.
%so that the encoder can produce equivalent sequence representations by minimizing it. 
Besides, we propose an agreement-based training scheme to learn universal mapping relationships for the translation directions with the same target language. 
%to improve the prediction consistency of the decoder based on the semantic-equivalent representations. 
We mixup the source and target sentences to obtain a pseudo sentence. Then, the decoder makes predictions separately conditioned on this pseudo sentence and the corresponding source or target sentences. 
We try to improve the prediction consistency by minimizing the KL divergence of the two output distributions. 
The experimental results on diverse multilingual datasets show that our method can bring 2\textasciitilde3 BLEU improvements over the strong baseline system and consistently outperform other contrast methods. The analysis proves that our method can better align the semantic space and improve the prediction consistency.

%minimize the discrepancy
%due to the difference of the sentence length and word order.  

%One of the ways to solve this problem is to make different languages share the same semantic space, that is, to make the sequence representations of sentences with the same meaning stay as close as possible. 

\section{Background}
In this section, we will give a brief introduction to the \textsc{Transformer}~\cite{VaswaniSPUJGKP17} model and the many-to-many multilingual translation. 
\subsection{The transformer}
We denote the input sequence of symbols as $\mathbf{x}=(x_1,\ldots,x_{nx})$ and the ground-truth sequence as $\mathbf{y}=(y_1,\ldots,y_{ny})$.
%and the translation as $\mathbf{y}^{*}=(y_1^{*},\ldots,y_{ny*}^{*})$. 
The transformer model is based on the encoder-decoder architecture. The encoder is composed of $\mathnormal{N}$ identical layers. Each layer has two sublayers. The first is a multi-head self-attention sublayer, and the second is a fully connected feed-forward network. Both of the sublayers are followed by a residual connection operation and a layer normalization operation. 
The input sequence $\mathbf{x}$ will be first converted to a sequence of vectors. %$\mathbf{E}_x=[E_x[x_1];\ldots;E_x[x_J]]$ where $E_x[x_j]$ is the sum of word embedding and position embedding of the source word $x_j$. 
Then, this sequence of vectors will be fed into the encoder, and the output of the $\mathnormal{N}$-th layer will be taken as source state sequences. We denote it as $\mathbf{H}_\mathbf{x}$. 
The decoder is also composed of $\mathnormal{N}$ identical layers. In addition to the same kind of two sublayers in each encoder layer, the cross-attention sublayer is inserted between them, which performs multi-head attention over the output of the encoder. 
%The final output of the $\mathnormal{N}$-th layer gives the target hidden states $\mathbf{S}=[\mathbf{s}_1;\ldots;\mathbf{s}_{K*}]$, where $\mathbf{s}_k$ is the hidden states of $y_k$. 
We can get the predicted probability of the $k$-th target word conditioned by the source sentence and the $k-1$ previous target words. The model is optimized by minimizing a cross-entropy loss of the ground-truth sequence with teacher forcing:
\begin{equation}\label{eq::loss}
    \mathcal{L}_{CE} = -\frac{1}{n_y} \sum_{k=1}^{n_y} \log p(y_k | \mathbf{y}_{<k}, \mathbf{x}; \theta),
\end{equation}
where $n_y$ is the length of the target sentence and $\theta$ denotes the model parameters. 

\subsection{Multilingual Translation}
We define $L=\{l_1, \ldots, l_M\}$ where $L$ is a collection of $M$ languages involved in the training phase. Following~\citet{JohnsonSLKWCTVW17}, we share all the model parameters for all the languages. Following~\citet{LiuGGLEGLZ20}, we add a particular language id token at the beginning of the source and target sentences, respectively, to indicate the language.

\begin{figure}[t!]
    \centering
    \includegraphics[width=1.0\columnwidth]{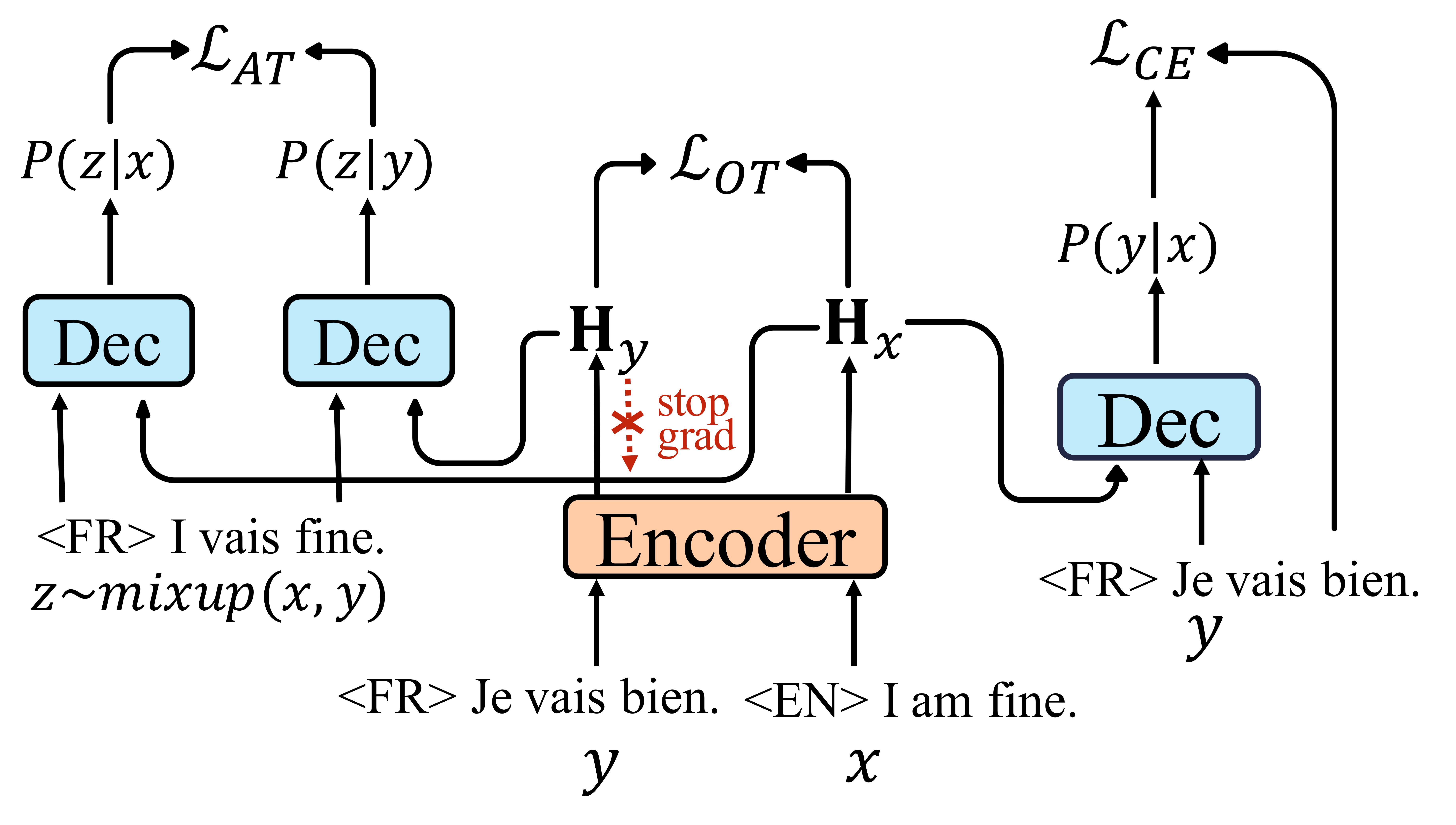}
    \caption{The training scheme of our method. $\mathbf{x}$ and $\mathbf{y}$ denote a pair of translations; $\mathbf{H_x}$ and $\mathbf{H_y}$ denote the corresponding state sequences. $\mathbf{z}$ is the pseudo sentence by mixuping $\mathbf{x}$ and $\mathbf{y}$. 'Dec' denotes the decoder and there is only one decoder in the model. 'stop-grad' denotes the stop-gradient operation during back propagation. $\mathcal{L}_{CE}$, $\mathcal{L}_{OT}$, and $\mathcal{L}_{AT}$ denote the cross entropy loss, optimal transport loss, and agreement-based training loss.}
    \label{fig:method}
\end{figure}

\section{Method}

The main idea of our method is to help the encoder output universal representations for all the languages and help the decoder map the semantic-equivalent representation from different languages to the target language's space.
%learn a universal semantic space for all the languages and 
%that the encoder, when fed with synonymous sentences, should produce equivalent state sequences that only carry semantic information. 
%So the decoder can generate the translation only based on the semantics, not the input language.
%Although the sequence representations always have differences in terms of word order length, etc., the decoder can still generate correct translations based on them, whatever the input language is.
%Only in this way can the knowledge learned on the seen language pairs be well transferred to the unseen language pairs. 
We propose two approaches to fulfill this goal. The first is to directly bridge the gap between the state sequences that carry the same semantics. The second is to force the decoder to make consistent predictions based on the semantic-equivalent sentences. Figure~\ref{fig:method} shows the overall training scheme.

%use the optimal transport theory to model the distance of state sequences and bridge the gap between them by minimizing this distance directly. The second is to force the decoder to make consistent predictions based on the equivalent state sequences. The overall training scheme is shown in Figure~\ref{fig:method}.

%bridge the gap between the sequence representations. The second is  

\subsection{Optimal Transport}
\textbf{Earth Mover's Distance} 
Based on the optimal transport theory~\cite{villani2009optimal,peyre2019computational}, the earth mover's distance (EMD) measures the minimum cost to transport the probability mass from one distribution to another distribution. 
%The more formal definition is as follows. 
Assuming that there are two probability distributions $\mu$ and $\mu'$, that are defined as:
\begin{equation}
\begin{split}
    &\mu = \{(\mathbf{w}_i, m_i) \}_{i=1}^{n}, \quad s.t. \sum_{i} m_i=1; \\
    &\mu' = \{(\mathbf{w}'_j, m'_j)\}_{j=1}^{n'}, \quad s.t. \sum_{j} m'_j=1,
% \quad  m_i \in [0, 1],
%\quad  m'_j \in [0, 1],
\end{split}
\end{equation}
where each data point $\mathbf{w}_i \in \mathbb{R}^d$ has a probability mass $m_i$ ($m_i>0$). There are $n$ data points in $\mu$. 
%There is also a transportation cost function $c$:
%\begin{equation}
%    c: \mathbb{R}^d \times \mathbb{R}^d \rightarrow \mathbb{R}.
%\end{equation}
We define a cost function $c(\mathbf{w}_i, \mathbf{w}'_j)$ that determines the cost of per unit between two points $\mathbf{w}_i$ and $\mathbf{w}'_i$. Given above, the EMD is defined as:
\begin{equation}
\begin{split}
    \mathcal{D}(\mu, \mu')&=\min_{\mathbf{T}\geq 0}\sum_{i,j} \mathbf{T}_{ij} c(\mathbf{w}_i,\mathbf{w'}_j) \\
    s.t. \quad & \sum_{j=1}^{n'} \mathbf{T}_{ij} = m_i, \forall i \in \{1, \ldots, n\}, \\
    & \sum_{i=1}^{n} \mathbf{T}_{ij} = m'_j, \forall j \in \{1, \ldots, n'\}.
\end{split}    
\end{equation}
$\mathbf{T}_{ij}$ denotes the mass transported from $\mu$ to $\mu'$.

\noindent \textbf{State Mover's Distance} Following EMD, we define the state mover's distance (SMD) to measure the minimum 'travel cost' between two state sequences. Given a pair of translations $\mathbf{x}=(x_1,\ldots,x_{nx})$, and $\mathbf{y}=(y_1,\ldots,y_{ny})$, we can get their corresponding state sequences after feeding them to the encoder, which are denoted as:
\begin{equation}
    \begin{split}
        & \mathbf{H}_\mathbf{x} = (\mathbf{h}_{1}, \ldots, \mathbf{h}_i, \ldots, \mathbf{h}_{nx}), \\
        & \mathbf{H}_\mathbf{y} = (\mathbf{h}'_{1}, \ldots, \mathbf{h}'_j, \ldots, \mathbf{h}'_{ny}),
    \end{split}
\end{equation}
where $nx$ and $ny$ denote the sentence length of the source and target sentences. We can regard $\mathbf{H_x}$ as a discrete distribution on the space $\mathbb{R}^d$, where the probability only occurs at each specific point $\mathbf{h}_i$. %Next, we need to define the specific amount of probability mass at each point. 
Next, several previous studies~\cite{SchakelW15,YokoiTASI20} have confirmed that the embedding norm is related to the word importance, and the important words have larger norms. 
Inspired by these findings, we also observe that the state vector has similar properties. The state vectors of essential words, such as content and medium-frequency words, have larger norms than unimportant ones, such as function words, high-frequency words. 
Therefore, we propose to use the normalized vector norm as the probability mass for each state point:
\begin{equation}
    m_i = \frac{|\mathbf{h}_i|}{\sum_{i}|\mathbf{h}_i|}, m'_j = \frac{|\mathbf{h}'_j|}{\sum_{j}|\mathbf{h}'_j|},
\end{equation}
where $|\cdot|$ denotes the norm of the vector.

Given above, we can convert the state sequences to distributions:
\begin{equation}
\begin{split}
   &\mu_{\mathbf{x}}^{\mathbf{H}}=\{(\mathbf{h}_i, \frac{|\mathbf{h}_i|}{\sum_{i}|\mathbf{h}_i|})\}_{i=1}^{nx}, \\ &\mu_{\mathbf{y}}^{\mathbf{H}}=\{(\mathbf{h}'_j, \frac{|\mathbf{h}'_j|}{\sum_{j}|\mathbf{h}'_j|})\}_{j=1}^{ny}.
\end{split}
\end{equation}
Then, the SMD is formally defined as follows:
\begin{equation}
\begin{split}
    \mathcal{D}&(\mu_{\mathbf{x}}^{\mathbf{H}}, \mu_{\mathbf{y}}^{\mathbf{H}})=\min_{\mathbf{T}\geq 0}\sum_{i,j} \mathbf{T}_{ij} c(\mathbf{h}_i,\mathbf{h'}_j), \\
    s.t. \quad & \sum_{j=1}^{ny} \mathbf{T}_{ij} = \frac{|\mathbf{h}_i|}{\sum_{i}|\mathbf{h}_i|}, \forall i \in \{1, \ldots, nx\}, \\
    & \sum_{i=1}^{nx} \mathbf{T}_{ij} = \frac{|\mathbf{h}'_j|}{\sum_{j}|\mathbf{h}'_j|}, \forall j \in \{1, \ldots, ny\}.
\end{split}    
\end{equation}

As illustrated before, we want decoder to make consistent predictions conditioned on the equivalent state sequences.
%the goal to bridge the gap between the state sequences is to help the decoder make consistent predictions based on them.
Considering that the vector norm and direction both have impacts on the cross-attention results of decoder, we use the Euclidean distance as the cost function. We didn't use the cosine similarity based metric, because it only considers the impact of vector direction. 
The proposed SMD is a fully unsupervised algorithm to align the contextual representations of the two semantic-equivalent sentences.
%To produce equivalent state sequences, we need to 

\noindent \textbf{Approximation of SMD} The exact computation to SMD is a linear programming problem with typical super $O(n^3)$ complexity, which will slow down the training speed greatly. 
We can obtain a relaxed bound of SMD by removing one of the two constraints, respectively.
%(removing both constraints results in the trivial lower bound $\mathbf{T}=0$.). 
Following~\citet{kusner2015word}, we remove the second constraints:
\begin{equation}
\begin{split}
    \mathcal{D}^{*}&(\mu_{\mathbf{x}}^{\mathbf{H}}, \mu_{\mathbf{y}}^{\mathbf{H}})=\min_{\mathbf{T}\geq 0}\sum_{i,j} \mathbf{T}_{ij} c(\mathbf{h}_i,\mathbf{h'}_j), \\
    s.t. \quad & \sum_{j=1}^{ny} \mathbf{T}_{ij} = \frac{|\mathbf{h}_i|}{\sum_{i}|\mathbf{h}_i|}, \forall i \in \{1, \ldots, nx\}. 
\end{split}    
\end{equation}
The above approximation must yield a lower bound to the exact SMD distance. The accurate SMD solution that satisfies both of the two constraints must also satisfy the first constraint.
Given the approximation, the optimal solution for each state vector $\mathbf{h}_i$ is to move all its probability mass to the most similar state vector $\mathbf{h}'_j$.  
Therefore, the approximation also enables the many-to-one alignment relationships during training. 
%In this way, we can get a reasonable approximation to the SMD with little extra training time.
We have also tried some approximation algorithms that can get a more accurate estimation of SMD, e.g., Sinkhorn algorithm\cite{cuturi2013sinkhorn}, IPOT~\cite{xie2020fast}. However, we have not observed consistent improvements in our preliminary experiments, and these algorithms also slow down the training speed significantly.

%However, these algorithms introduce more hyperparameters and also take more time, because they all depend on iteration. We also have not observed consistent performance improvements in our preliminary experiments after replacing the above approximation with these algorithms. 

\noindent \textbf{Objective Function} 
%We want to minimize the SMD of both sides, so we define a symmetrical loss as:
We define a symmetrical loss to minimize the SMD of both sides:
\begin{equation}\label{eq:ot}
    \mathcal{L}_{OT} = \frac{1}{2}\left(\mathcal{D}^{*}(\mu_{\mathbf{x}}^{\mathbf{H}}, \mu_{\mathbf{y}}^{\mathbf{H}}) + \mathcal{D}^{*}(\mu_{\mathbf{y}}^{\mathbf{H}}, \mu_{\mathbf{x}}^{\mathbf{H}})\right).
\end{equation}

\subsection{Agreement-based Training} 
\textbf{Theoretical Analysis}
In zero-shot translation, the decoder should map the semantic representations from different languages to the target language space, even if it has never seen the translation directions during training. This ability needs the model to make consistent predictions based on the semantic-equivalent sentences, whatever the input language is.
%Then, the decoder can generate the correct translation based on the semantics carried by the state sequences, even if it has never seen the translation directions during050training.
%We need to make 
%The above method only focus on the encoder, and can directly reduce the difference between the equivalent state sequences.
%In addition to directly reducing the difference between the equivalent state sequences, 
To improve the prediction consistency of the model, we propose an agreement-based training method.
%which force the decoder to make consistent predictions based on them. 
Because the source sentence $\mathbf{x}$ and target sentence $\mathbf{y}$ are semantically equivalent, the probability of predicting any other sentence $\mathbf{z}$ based on them should be always equal theoretically, which is denoted as:
\begin{equation}\label{eq:at1}
    p(\mathbf{z}|\mathbf{x}) = p(\mathbf{z}|\mathbf{y}).
\end{equation}
%More
Specifically, the predicted probabilities of the $k$-th target word conditioned by the first $k-1$ words of $\mathbf{z}$ and the source and target sentences is equal:
\begin{equation}\label{eq:at2}
    p(z_k|\mathbf{z}_{<k}, \mathbf{x}; \theta) = p(z_k|\mathbf{z}_{<k}, \mathbf{y}; \theta),
\end{equation}
where $\theta$ denotes the model parameters. Optimizing Equation~\ref{eq:at2} can not only help the encoder produce universal semantic representations but also help the decoder map different source languages to the particular target language space indicated by $\mathbf{z}$.

%ignore the spurious relations between the input and output language and focus more on the semantics of the input sentence. 
\noindent \textbf{Mixup for $\mathbf{z}$}
Although Equation~\ref{eq:at2} is theoretically attractive, the choice of sentence $\mathbf{z}$ has a significant influence on the above optimization. If we use a random sentence as $\mathbf{z}$, which is not related to $\mathbf{x}$ and $\mathbf{y}$, the prediction makes no sense, and the model learns helpful nothing.
%makes optimization difficult.
%the potential prediction space will be too large to make optimization difficult.
If we use either $\mathbf{x}$ or $\mathbf{y}$ directly, this will cause information leakage on one side of Equation~\ref{eq:at2}. As a result, the prediction difficulty between the two sides differs significantly, and it is hard for one side to catch up with the other side. 
Given the above, we need a inter-sentence that is "between" $\mathbf{x}$ and $\mathbf{y}$.
Inspired by the success of mixup technique in NLP~\cite{ZhangYZ20,ChengWJM21}, we generate a pseudo sentence by hard mixuping $\mathbf{x}$ and $\mathbf{y}$ at token-level. We truncate the longer sentences 
of $\mathbf{x}$ and $\mathbf{y}$ to make them equal in length. Since these two sentences are translation pairs, their sentence lengths are usually close, truncating will not significantly reduce the length of the longer sentence and will not enhance the decoder learn shorter outputs. We denote the truncated sentence as $\mathbf{x}'$ and $\mathbf{y}'$, and their length as $n'$. Then we can generate $\mathbf{z}$ as:
\begin{equation}
    \mathbf{z}= \mathbf{g} \odot \mathbf{x}' + (1-\mathbf{g}) \odot \mathbf{y}',
\end{equation}
where $\mathbf{g} \in \{0, 1\}^{n'}$, $\odot$ denotes the element-wise product. Each element in $\mathbf{g}$ is sampled from Bernoulli$(\lambda)$, where the parameter $\lambda$ is sampled from Beta$(\alpha, \beta)$, and $\alpha$ and $\beta$ are two hyperparameters. The language tag in $\mathbf{z}$, which determines the translation direction, is either come from $\mathbf{x}$ or $\mathbf{y}$. 

\noindent \textbf{Objective Function} Similar to Equation~\ref{eq:ot}, we define another symmetrical loss based on the KL divergence of the model prediction distributions:
\begin{equation}\label{eq:kl}
\begin{split}
    \mathcal{L}_{AT} =& \frac{1}{2 n'}\sum_{k=1}^{n'} KL\left(p(z_k|\mathbf{z}_{<k}, \mathbf{H_x}) || p(z_k|\mathbf{z}_{<k}, \mathbf{H_y})\right)  \\
    & + KL\left(p(z_k|\mathbf{z}_{<k}, \mathbf{H_y}) || p(z_k|\mathbf{z}_{<k}, \mathbf{H_x})\right). 
\end{split}
\end{equation}
We omit the model parameters for convenience. 

\subsection{The Final Loss}
The final loss consists of three parts, the cross entropy loss (Equation~\ref{eq::loss}), the optimal transport loss based on SMD (Equation~\ref{eq:ot}) and the KL divergence loss for the agreement-based training (Equation~\ref{eq:kl}):
\begin{equation}\label{eq:allloss}
    \mathcal{L} = \mathcal{L}_{CE} + \gamma_1 |\mathbf{x}| \mathcal{L}_{OT} + \gamma_2 \mathcal{L}_{AT}
\end{equation}
where $\gamma_1$ and $\gamma_2$ are two hyperparameters that control the contributions of the two regularization loss terms. Since $\mathcal{L}_{OT}$ is calculated on the sentence-level and the other two losses are calculated on the token-level, we multiply the averaged sequence length $|\mathbf{x}|$ to $\mathcal{L}_{OT}$.
Among these three losses, the first term dominates the parameter update of the model, and determines the model performance mostly.
The latter two regularization loss terms only slightly modify the directions of the gradients.
Because the first loss term does not depend on $\mathbf{H_y}$, we apply the stop-gradient operation to $\mathbf{H_y}$ (Figure~\ref{fig:method}), which means that the gradients will not pass through $\mathbf{H_y}$ to the encoder.
%and only the gradient produced by the latter two losses does not necessarily improve the translation quality. Therefore, we apply the stop-gradient operation to $\mathbf{H_y}$ (Figure~\ref{fig:method}), which means that the gradients will not pass through $\mathbf{H_y}$ to the encoder.

%the distance between the two sides cannot be effectively shortened.

%\noindent \textbf{z\textasciitilde mixup(x,y)}

%can also force the deocder to produce a consistent output based on them.

%Precisely, an optimal $\mathbf{T}^{*}$ matrix is defined as:

%(satisfying both constraints) must remain a feasible solution if one constraint is removed.

\begin{table}[t]
\centering
\resizebox{\columnwidth}!{
\begin{tabular}{l|l|c}
\hline
Dataset & Language Pairs                & Size             \\ \hline
IWSLT   & En$\leftrightarrow$\{De, It, Nl, Ro\}         & 1.79M            \\
IWSLT-b & Nl$\leftrightarrow$De$\leftrightarrow$En$\leftrightarrow$It$\leftrightarrow$Ro                & 1.79M            \\
PC-6    & En$\leftrightarrow$\{Kk, Tr, Ro, Cs, Ru\}    & 7.9M \\
OPUS-7  & En$\leftrightarrow$\{De, Fr, Nl, Ru, Zh, Ar\} & 11.6M               \\ \hline
\end{tabular}
}
\caption{ The statics of our datasets.}
% The column of size shows the data amount of each language pair. }

\label{tab:data}
\end{table}

\section{Experiments}
\subsection{Data Preparation}
We conduct experiments on the following multilingual datasets: IWSLT17, PC-6, and OPUS-7. 
The brief statistics of the training set are in Table~\ref{tab:data}. We put more details in the appendix.

\noindent \textbf{IWSLT17}~\cite{cettolo2017overview} %This dataset contains about 1.79M sentence pairs.
We simulate two scenarios. The first (IWSLT) is English-pivot, where we only retain the parallel sentences from/to English. The second (IWSLT-b) has a chain of pivots, where two languages are connected by a chain of pivot languages. Each translation direction has about 0.22M sentence pairs. Both of the two scenarios have eight supervised translation directions and twelve zero-shot translation directions. We use the official validation and test sets. 

\noindent \textbf{PC-6} 
The PC-6 dataset is extracted from the PC-32 corpus~\cite{LinPWQFZL20}. The data amount of different language pairs is unbalanced, ranging from 0.12M to 1.84M. 
%just as multilingual translation is in many practical application scenarios. 
This dataset has ten supervised and twenty zero-shot translation directions. We use the validation and test sets collected from WMT16\textasciitilde19 for the supervised directions. The zero-shot validation and test sets are extracted from the WikiMatrix~\cite{SchwenkCSGG21}, each containing about 1K\textasciitilde2K sentences pairs. 

\noindent \textbf{OPUS-7} 
The OPUS-7 dataset is extracted from the OPUS-100 corpus~\cite{ZhangWTS20}. The language pairs come from different language families and have significant differences. This dataset has twelve supervised translation directions and thirty zero-shot translation directions. We use the standard validation and test sets released by~\citet{ZhangWTS20}. We concatenate the zero-shot test sets with the same target language for convenience.  

We use the Stanford word segmenter~\cite{TsengCAJM05,MonroeGM14} to segment Arabic and Chinese, and the Moses toolkit~\cite{KoehnHBCFBCSMZDBCH07} to tokenize other languages. 
Besides, integrating operations of 32K is performed to learn BPE~\cite{SennrichHB16a}.

% \usepackage{multirow}
\begin{table*}[t]
\centering
\subtable{
\resizebox{2.03\columnwidth}!{
\begin{tabular}{c|ccccccccccccc|c}
\hline
IWSLT & \multicolumn{2}{c}{De-It} & \multicolumn{2}{c}{De-Nl} & \multicolumn{2}{c}{De-Ro} & \multicolumn{2}{c}{It-Ro} & \multicolumn{2}{c}{It-Nl} & \multicolumn{2}{c}{Nl-Ro} & \multirow{2}{*}{\begin{tabular}[c]{@{}c@{}}Zero\\ Avg.\end{tabular}} & \multirow{2}{*}{\begin{tabular}[c]{@{}c@{}}Sup.\\ Avg\end{tabular}} \\ \cline{1-1}
Model & $\rightarrow$ & $\leftarrow$ & $\rightarrow$ & $\leftarrow$ & $\rightarrow$ & $\leftarrow$ & $\rightarrow$ & $\leftarrow$ & $\rightarrow$ & $\leftarrow$ & $\rightarrow$ & $\leftarrow$ &  &  \\ \hline
ZS & 15.64 & 15.28 & 18.46 & 18.14 & 14.42 & 14.98 & 17.91 & 20.14 & 18.16 & 18.79 & 15.81 & 16.41 & 17.01 & 30.62 \\
SRA & 16.44 & 16.45 & 18.44 & 19.15 & 15.07 & 15.83 & 18.52 & 21.52 & 19.3 & 19.1 & 16.83 & 17.66 & 17.85 & 30.41 \\
SF & 16.34&	15.77&	18.37&	18.16&	14.74&	15.25&	18.54&	21.64&	18.6&	19.18&	16.09&	16.94&	17.46&	30.5 \\
CL & 17.37 & 16.58 & 19.69 & 19.5 & 15.51 & 16.25 & 18.91 & 22.58 & 18.78 & 20.02 & 17.27 & 17.91 & 18.36 & 30.39 \\
DisPos & 16.62 & 15.64 & 19.64 & 18.78 & 15.07 & 15.96 & 18.67 & 21.56 & 19.01 & 20.15 & 16.46 & 18.18 & 17.97 & 30.49 \\ 
DT & 16.82 & 15.81 & 18.74 & 18.64 & 15.12 & 16.32 & 18.70 & 22.13 & 18.92 & 19.29 & 16.21 & 18.22 & 17.91 & 30.51 \\
%TLP & 16.38 & 15.72 & 18.42 & 18.2 & 14.59 & 15.39 & 18.61 & 21.38 & 18.77 & 19.88 & 16.11 & 17.01 & 17.54 & 30.5 \\
TGP & 16.77 & \textbf{18.51} & 14.58 & 17.12 & \textbf{16.84} & \textbf{16.88}  & 19.42 & 19.25 & 20.01 & 19.04 & 21.67 & 18.43 & 18.21 & \textbf{30.66} \\
LMP & 16.87 & 18.44 & 15.05 & 16.66 & 16.20 & 16.12 & 19.04 & 19.05 & 19.35 & 18.68 & \textbf{22.17} & 17.97 & 17.96 & 30.52 \\ 
\hdashline
PivT & 18.31 & 17.9 & 19.99 & 19.33 & 15.54 & 17.45 & 19.77 & 22.97 & 21.43 & 21.44 & 17.57 & 19.82 & 19.29 & - \\ \hdashline
ZS+OT & 17.35 & 17.08 & 19.77 & 19.05 & 15.66 & 16.17 & \textbf{19.71} & 22.32 & \textbf{20.18} & \textbf{20.57} & 16.87 & 18.09 & 18.56 & 30.42 \\
ZS+AT & 16.37 & 15.84 & 19.11 & 18.41 & 14.85 & 15.59 & 18.37 & 21.09 & 18.77 & 19.4 & 15.86 & 17.46 & 17.59 & 30.55 \\
Ours & \textbf{17.53} & 17.03 & \textbf{19.94} & \textbf{19.67} & 15.61 & 16.57 & 19.23 & \textbf{22.42} & 20.05 & 20.23 & \textbf{17.05} & \textbf{18.64} & \textbf{18.66} & 30.52 \\ \hline
\end{tabular}
}
}

\subtable{
\resizebox{2.03\columnwidth}!{
\begin{tabular}{c|ccccccccccccc|c}
\hline
IWSLT-b & \multicolumn{2}{c}{De-It} & \multicolumn{2}{c}{En-Nl} & \multicolumn{2}{c}{De-Ro} & \multicolumn{2}{c}{En-Ro} & \multicolumn{2}{c}{It-Nl} & \multicolumn{2}{c}{Nl-Ro} & \multirow{2}{*}{\begin{tabular}[c]{@{}c@{}}Zero\\ Avg.\end{tabular}} & \multirow{2}{*}{\begin{tabular}[c]{@{}c@{}}Sup.\\ Avg.\end{tabular}} \\ \cline{1-1}
Model & $\rightarrow$ & $\leftarrow$ & $\rightarrow$ & $\leftarrow$ & $\rightarrow$ & $\leftarrow$ & $\rightarrow$ & $\leftarrow$ & $\rightarrow$ & $\leftarrow$ & $\rightarrow$ & $\leftarrow$ &  &  \\ \hline
ZS & 17.79 & 17.3 & 25.48 & 30.99 & 15.65 & 17.28 & 21.7 & 30.14 & 20.79 & 21.02 & 15.74 & 17.28 & 20.93 & \textbf{30.46} \\
SRA & 18.09 & 18.05 & 26.52 & 31.15 & 15.8 & 17.43 & 22.24 & 30.19 & 20.35 & 20.65 & 16.39 & 17.83 & 21.22 & 30.29 \\
SF& 18.25&	17.61&	26&	31.28&	16.06&	17.51&	22.43&	30.51&	20.67&	20.82&	16.2&	17.24&	21.21&	30.35 \\
CL & \textbf{18.49} & 18.29 & 26.88 & 31.46 & 15.71 & 17.23 & 23.01 & 30.78 & 20.62 & 20.8 & 16.58 & 18.17 & 21.5 & 30.28 \\
DisPos & 17.98 & 17.35 & 26.26 & 31.13 & 15.75 & \textbf{18.07} & 22.95 & 30.45 & \textbf{21.02} & 20.58 & 16.38 & 18.28 & 21.35 & 29.89 \\ 
%TLP & 18.36 & 18.02 & 26.33 & 31.03 & 15.89 & 17.75 & 22.33 & 30.39 & 20.65 & 20.59 & 16.31 & 17.51 & 21.26 & 30.29 \\ 
TGP & 18.22 & 18.69 & 26.62 & 30.96 & 15.57 & 17.26 & 23.21 & 30.22 & 20.62 & 20.38 & 16.58 & 17.65 & 21.33 & 30.33 \\
LMP & 18.36 & \textbf{18.83} & 27.2 & 30.5 & 16.05 & 17.05 & \textbf{23.99} & 29.38 & 20.57 & 19.83 & 16.72 & 17.56 & 21.33 & 30.37 \\
\hdashline
PivT & 18.38 & 19.08 & 27.3 & 28.02 & 15 & 16.35 & 23.72 & 28.72 & 20.34 & 19.45 & 15.7 & 16.8 & 20.74 & - \\ \hdashline
ZS+OT & 18.09 & 18.06 & 26.6 & \textbf{31.69} & 15.76 & 17.19 & 23.46 & \textbf{30.99} & 20.31 & \textbf{20.86} & 16.92 & 18.05 & 21.49 & 30.37 \\
ZS+AT & 18.23 & 17.51 & 26.24 & 31.12 & \textbf{16.19} & 17.5 & 22.64 & 30.33 & 20.72 & 20.59 & 16.29 & 17.64 & 21.25 & 30.39 \\
Ours & 18.41 & 18.05 & \textbf{27.39} & 31.36 & 16.15 & 17.48 & 23.22 & 30.9 & 20.68 & 20.82 & \textbf{17.03} & \textbf{18.29} & \textbf{21.64} & 30.33 \\ \hline
\end{tabular}
}
}

\subtable{
\resizebox{2.03\columnwidth}!{
\begin{tabular}{c|ccccccc||c|cccccccc}
\hline
PC-6 & x$\rightarrow$Kk & x$\rightarrow$Tr & x$\rightarrow$Ro & x$\rightarrow$Cs & x$\rightarrow$Ru & \begin{tabular}[c]{@{}c@{}}Zero \\ Avg.\end{tabular} & \begin{tabular}[c]{@{}c@{}}Sup. \\ Avg.\end{tabular} & OPUS-7 & x$\rightarrow$De & x$\rightarrow$Fr & x$\rightarrow$Nl &x$\rightarrow$Ru & x$\rightarrow$Zh & x$\rightarrow$Ar & \begin{tabular}[c]{@{}c@{}}Zero \\ Avg.\end{tabular} & \begin{tabular}[c]{@{}c@{}}Sup. \\ Avg.\end{tabular} \\ \hline
ZS & 5.87 & 9.29 & 14.23 & 13.55 & 16.83 & 11.95 & \textbf{21.73} & ZS & 13.58 & 22.63 & 17.96 & 15.42 & 29.78 & 21.58 & 20.15 & \textbf{34.2} \\
SRA & 5.90 & 10.09 & 17.36 & 15.85 & 19.31 & 13.68 & 21.66 & SRA & 17.04 & 26.12 & 19.29 & 20.9 & 31.99 & 22.01 & 22.89 & 33.97 \\
SF& 4.76&	9.95&	17.77&	15.83&	20.10&	13.68&	21.64& SF& 15.99	&25.2&	18.2&	20.85&	31.65&	21.5&	22.23&	33.99 \\
CL & 6.07 & 10.72 & 17.96 & 16.14 & 21.58 & 14.49 & 21.54 & CL & 17.41 & 26.19 & 19.66 & 21.1 & 32.52 & 21.69 & 23.09 & 33.86 \\
DisPos & 6.60 & 10.14 & 15.47 & 15.89 & 18.70 & 12.51 & 21.45 & DisPos & 15.95 & 25.36 & 18.86 & 19.75 & 31.34 & 22.08 & 22.22 & 34.12 \\ 
DT & 6.92 & 10.49 & 17.37 & 15.63 & 21.74 & 14.43 & 21.61 & DT & 14.97 & 23.95 & 18.10 & 18.91 & 29.65 & 20.68 & 21.04 & 34.03 \\ 
%TLP & 5.25 & 10.22 & 17.71 & 15.52 & 20.44 & 13.83 & 21.61 & TLP & 16.03 & 24.75 & 18.55 & 20.03 & 31.58 & 20.96 & 21.98 & 34.05 \\
TGP & \textbf{7.33} & 10.98 & \textbf{20.63} & 13.81 & 21.21 & 14.79 & 21.58 & TGP & 16.86 & 25.65 & 18.99 & 20.83 & 32.47 & 21.47 & 22.71 & 34.18 \\
LMP & 4.45 & 8.50 & 16.42 & 15.25 & 19.28 & 12.78 & 21.71 & LMP & 14.65 & 23.94 & 18.36 & 19.02 & 30.58 & 20.99 & 21.26 & 34.07 \\
\hdashline
PivT & 4.29 & 10.59 & 19.23 & 17.22 & 21.65 & 14.58 & - & PivT & 17.97 & 28.37 & 19.76 & 22.97 & 34.08 & 23.74 & 24.48 & - \\ \hdashline
ZS+OT & 6.22 & 11.08 & 18.74 & 16.86 & 22.61 & 15.1 & 21.6 & ZS+OT & 17.56 & 26.70 & 19.54 & 21.88 & 32.42 & 22.48 & 23.43 & 34.02 \\
ZS+AT & 6.04 & 10.74 & 17.92 & 15.69 & 20.63 & 14.2 & 21.72 & ZS+AT & 16.78 & 25.89 & 18.93 & 21.21 & 32.02 & 21.72 & 22.75 & 34.1 \\
Ours & 6.58 & \textbf{11.44} & 18.55 & \textbf{17.11} & \textbf{22.77} & \textbf{15.29} & 21.68 & Ours & \textbf{17.60} & \textbf{26.74} & \textbf{19.68} & \textbf{21.91} & \textbf{32.63} & \textbf{23.24} & \textbf{23.63} & 34.17 \\ \hline
\end{tabular}
}
}

\caption{The overall BLEU scores on the test sets. "Zero Avg." and "Sup. Avg." denote the average BLEU scores on the zero-shot and supervised directions. The "x" in the third table denotes all languages except for the target language. The highest scores are marked in bold for all models except for the "PivT" system in each column.}
\label{tab:res}
\end{table*}

\subsection{Systems} 
We use the open-source toolkit called {\em Fairseq-py}~\cite{OttEBFGNGA19} as our Transformer system. We implement the following systems:

\noindent \textbullet \ \textbf{Zero-Shot (ZS)} The baseline system which is trained only with the cross-entropy loss (Equation~\ref{eq::loss}). Then the model is tested directly on the zero-shot test sets.

\noindent \textbullet \ \textbf{Pivot Translation (PivT)}~\cite{ChengYLSX17} The same translation model as ZS. The model first translates the source language to the pivot language and then generates the target language. 

\noindent \textbullet \textbf{Sentence Representation Alignment (SRA)}~\cite{abs-1903-07091} This methods adds an regularization loss to minimize the discrepancy of the source and target sentence representations.
%We vary the hyperparameter of the regularization loss from $0.1$ to $1$ to tune the performance.
\begin{equation}
    \mathcal{L} = \mathcal{L}_{CE} + \gamma Dis(Enc(s), Enc(t)),
\end{equation}
where 'Dis' denotes the distance function and 'Enc($\cdot$)' denotes the sentence representations. We use the averaged sentence representation and Euclidean distance function because we find they work better. 
We vary the hyperparameter $\gamma$ from $0.1$ to $1$ to tune the performance.   

\noindent \textbullet \ \textbf{Softmax Forcing (SF)}~\cite{PhamNHW19} This method enable the decoder to generate the target sentence from itself by adding an extra loss:
\begin{equation}
    \mathcal{L}_{SF}=\gamma \sum_k^{n_y}KL(p(y_k|\mathbf{y}_{<k},\mathbf{x})||p(y_k|\mathbf{y}_{<k},\mathbf{y}))
\end{equation}
The $\gamma$ is tuned as in the 'SRA' system.

\noindent \textbullet \ \textbf{Contrastive Learning (CL)}~\cite{PanWWL20} This method adds an extra contrastive loss to minimize the representation gap of similar sentences and maximize that of irrelevant sentences:
\begin{equation}
    \mathcal{L}_{CL}=-\gamma \log \frac{e^{sim^{+}(\mathcal{R}(s),\mathcal{R}(t))/\tau}}{\sum_w e^{sim^{-}(\mathcal{R}(s),\mathcal{R}(w))/\tau}},
\end{equation}
where $+$ and $-$ denote positive and negative sample pairs, $\mathcal{R}(\cdot)$ denotes the averaged state representations. We set $\tau$ as 0.1 as suggested in the paper and tune $\gamma$ as in the 'SRA' system. 

\noindent \textbullet \ \textbf{Disentangling Positional Information (DisPos)}~\cite{LiuNCGL20} This method removes the residual connections in a middle layer of the encoder to get the language-agnostic representations.

\noindent \textbullet \ \textbf{Denosing Training (DT)}~\cite{abs-2109-04705} This method introduces a denoising auto-encoder objective during training. 

%\noindent \textbullet \ \textbf{Target Language Prediction (TLP)}~\cite{abs-2109-04778} This method leverages an auxiliary language prediction task during training.   

\noindent \textbullet \ \textbf{Target Gradient Projection (TGP)}~\cite{abs-2109-04778} This method projects the training gradient to not conflict with the oracle gradient of a small amount of direct data.

\noindent \textbullet \ \textbf{Language Model Pre-training (LMP)}~\cite{GuWCL19} This method strengthens the decoder language model prior to machine translation training. 
%\noindent 

The following systems are implemented based on our method:

\noindent \textbullet \ \textbf{ZS+OT} 
We only add the optimal transport loss (Equation~\ref{eq:ot}) during training. We vary the hyperparameter $\gamma_1$ from $0.1$ to $1$, and we find that it can constantly improve the performance whatever $\gamma_1$ is. The detailed results and the final setting about the hyperparameter are put in the appendix. 

\noindent \textbullet \ \textbf{ZS+AT}  We only add the agreement-based training loss (Equation~\ref{eq:kl}) during training. The $\alpha$ and $\beta$ in the beta distribution are set as $6$ and $3$, respectively. We vary the hyperparameter $\gamma_2$ from ${10}^{-4}$ to $0.1$. 

\noindent \textbullet \ \textbf{ZS+OT+AT (Ours)} The model is trained with the complete objective function (Equation~\ref{eq:allloss}). The hyperparameters are set according to the searched results of the above two systems and are listed in the appendix. 

\noindent \textbf{Implementation Details} All the systems are implemented as the base model configuration in~\citet{VaswaniSPUJGKP17} strictly.  
We employ the Adam optimizer with $\beta_1 = 0.9$ and $\beta_2 = 0.98$. 
We use the inverse square root learning scheduler and set the $warmup\_steps=4000$ and $lr=0.0007$. We set dropout as 0.3 for the IWSLT datasets and 0.1 for the for the PC-6 and OPUS-7 datasets. 
%Since the IWSLT17 dataset is smaller than the other two datasets, we set the dropout rate as 0.3 for better performance. 
All the systems are trained on 4 RTX3090 GPUs with the update frequency 2. The max token is 4096 for each GPU.
For the IWSLT data sets, we first pretrain the model with the cross-entropy loss (Equation~\ref{eq::loss}) for 20K steps and then continually train the model combined with the proposed loss terms for 80K steps. For the PC-6 and OPUS-7 datasets, the pre-training steps and continual-training steps are both 100k.

%we first pretrain the model for 100K steps and then continual-train the model with the proposed loss terms for 200K steps. 
%For SF, SRA, CL, and our systems, we first pretrain the model only with the cross-entropy loss (Equation~\ref{eq::loss}). Then we add the corresponding regularization losses in the middle stage of training and train the model until it converges.
%because we find it can save training time and improve the model performance. %We vary the hyperparameter $\gamma$ from $0.1$ to $1$ to tune the performance for the SF, SRA, and CL systems.

\begin{figure*}[t]
    \centering
    \subfigure[ZS]{
        \includegraphics[width=0.48\columnwidth]{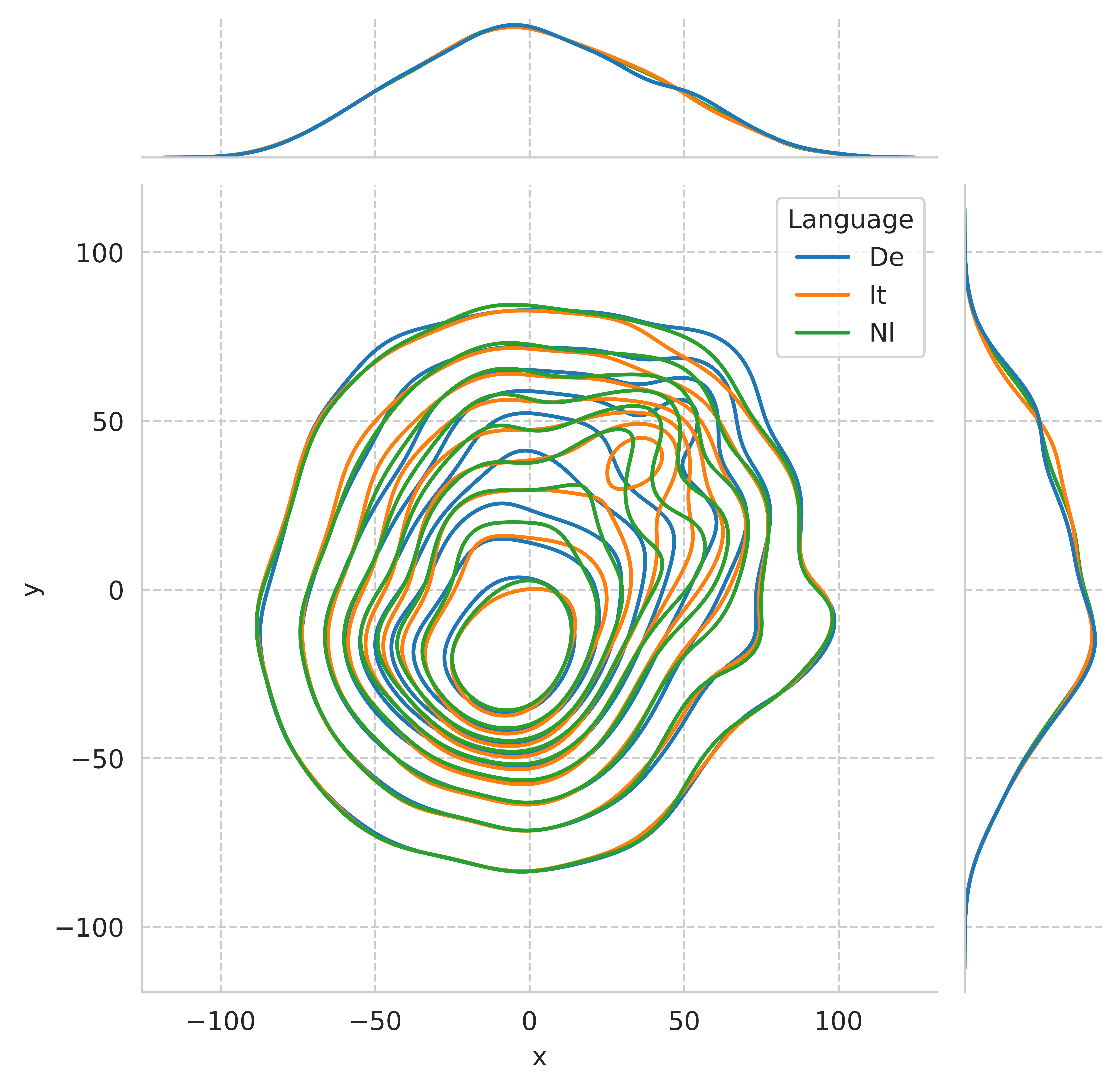}
    }
    \subfigure[SRA]{
        \includegraphics[width=0.48\columnwidth]{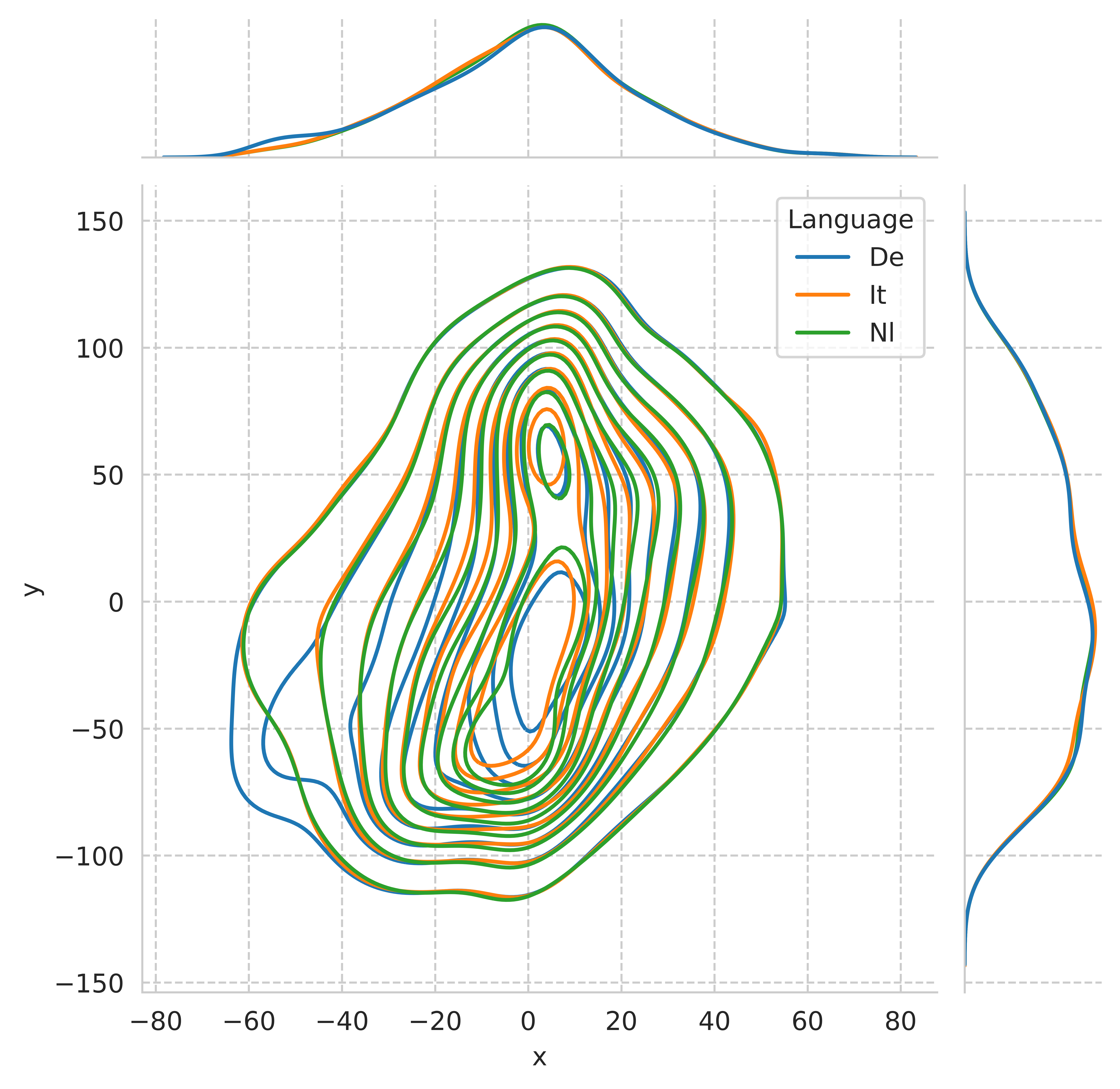}
    }
    \subfigure[CL]{
        \includegraphics[width=0.48\columnwidth]{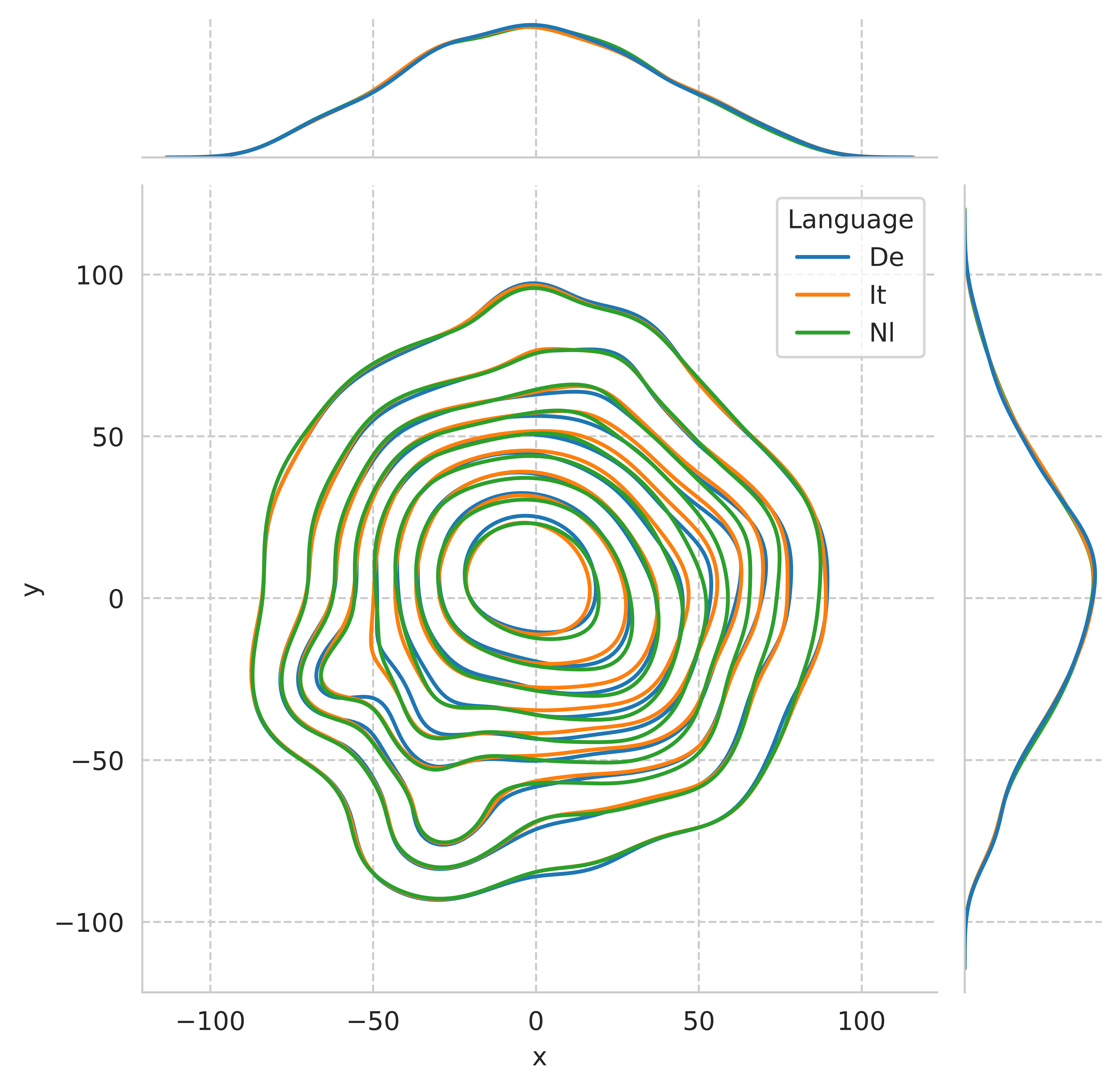}
    }
    \subfigure[Ours]{
        \includegraphics[width=0.48\columnwidth]{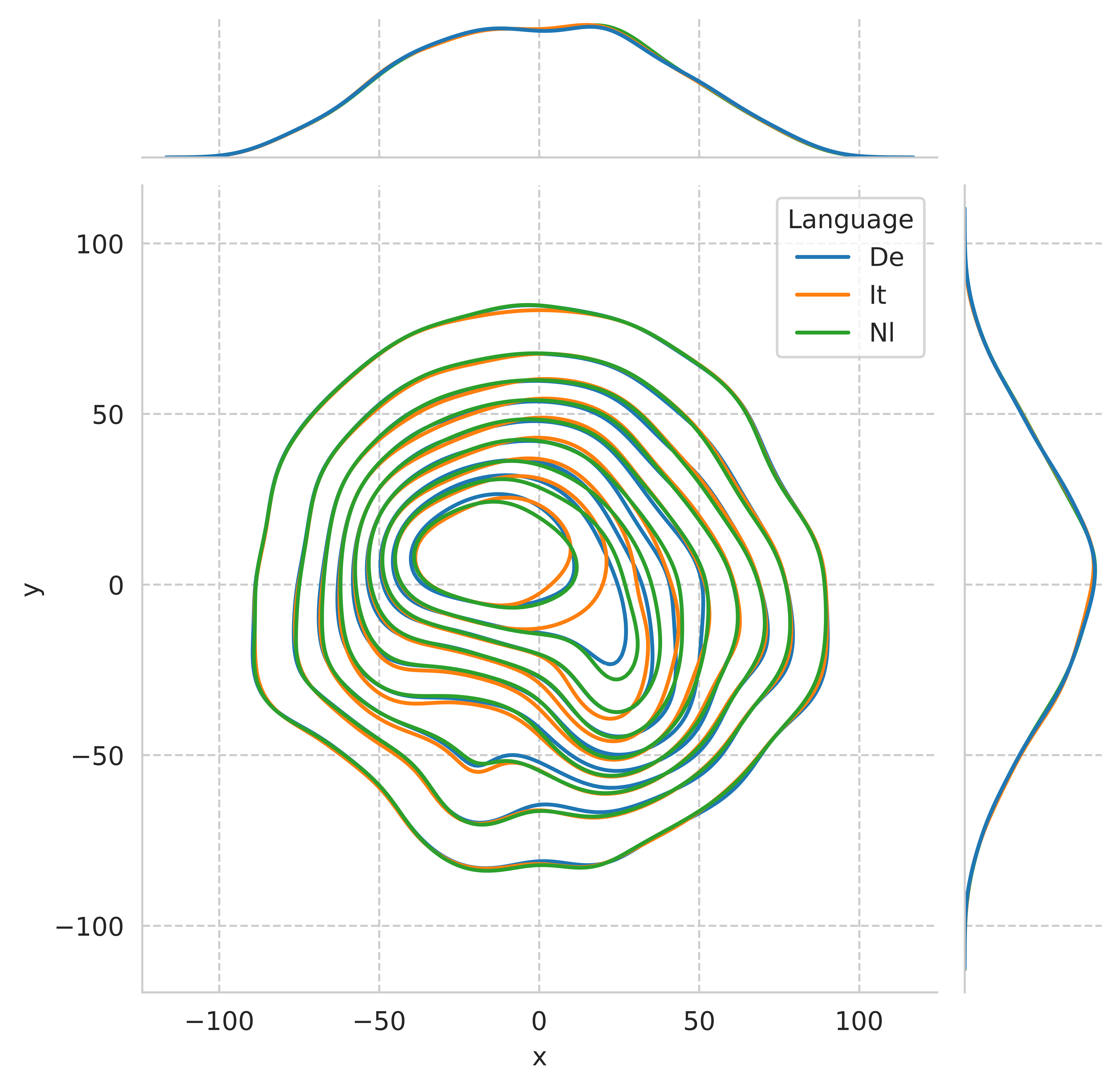}
    }
    \caption{The visualization of sentence representation after dimension reduction on the IWSLT three-way-parallel test sets. The blue line denotes Germany, the orange line denotes Italian, and the green line denotes Dutch.}
    \label{fig:vis}
\end{figure*}

\subsection{Main Results}
All the results (including the intermediate results of the 'PivT' system) are generated with beam size = $5$ and length penalty $\alpha = 0.6$.
The translation quality is evaluated using the case-sensitive BLEU~\cite{PapineniRWZ02} with the {\em SacreBLEU} tool~\cite{post-2018-call}. We report the tokenized BLEU for Arabic, char-based BLEU for Chinese, and detokenized BLEU for other languages\footnote{BLEU+case.mixed+numrefs.1+smooth.exp+\\tok.\{13a,none,zh\}+version.1.5.1}. The main results are shown in Table~\ref{tab:res}. We report the averaged BLEU with the same target language on the PC-6 and OPUS-7 dataset for display convenience, and the detailed results are in the appendix. The 'Ours' system significantly improves over the 'ZS' baseline system and outperforms other zero-shot-based systems on all datasets. The two proposed methods, OT and AT, can both help the model learn universal and cross mappings , so they both can improve the model performance independently.
These two methods also complement each other and can further improve the performance when combined together. Besides, 'Ours' system can even exceed the 'PivT' system when the distant language pairs in the IWSLT-b or the low-resource language pairs in the PC-6 bring severe error accumulation problems. We also compare the training speed and put the results in the appendix.

\begin{table}[t]
\centering
\begin{tabular}{c|cccc}
\hline 
IWSLT & x-De & x-It & x-Nl & Avg. \\ \hline \hline
ZS & 21.5 & 20.79 & 19.99 & 20.76 \\
SRA & 21.79 & 21.92 & 20.67 & 21.46 \\
CL & 23.47 & 21.52 & 21.09 & 22.03 \\
Ours & \textbf{23.6} & \textbf{23.33} & \textbf{21.48} & \textbf{22.80} \\ \hline
\end{tabular}
\caption{The pair-wise BLEU on the IWSLT three-way-parallel test sets.}
\label{tab:pbleu}
\end{table}

\section{Analysis}
In this section, we try to understand how our method improves the zero-shot translation. 
%and what improvements our method brings. 

\subsection{Sentence Representation Visualization}
To verify whether our method can better align different languages' semantic space, we visualize each model's encoder output with the IWSLT test sets. We first select three languages: Germany, Italian, and Dutch. Then we filter out the overlapped sentences of the three languages from the corresponding test sets and create a new three-way-parallel test set. Next, we feed all the sentences to the encoder of each model and average the encoder output to get the sentence representation. Last, we apply dimension reduction to the representation with t-SNE~\cite{van2008visualizing}. The visualization result in Figure~\ref{fig:vis}(a) shows that the 'ZS' system cannot align the three languages well, which partly confirms our assumption that the conventional MNMT cannot learn universal representations for all languages.
As a contrast, the 'Ours' system (d) can draw the representation closer and achieve comparative results as the 'CL' system (c) without large amounts of negative instances to contrast. 
The visualization results confirm that our method can learn good universal representation for different languages.

%Meanwhile, the 'SRA' system \(b\) can align the representations better, and the 'CL' system \(c\) can further draw the presentation closer because of using negative instances for contrast.
%Because of using negative instances for contrast, the 'CL' system can further draw the presentation closer. 
%Although our method reduces the gap on the token level, the figure \(d\) shows our method can learn good universal representation for different languages.

\subsection{Inspecting Prediction Consistency}
To verify whether our method 
%can help learn universal cross-mapping relations for all the languages,
can help map the semantic representation from different languages to the same space of the target language, 
we inspect the prediction consistency of the models when the model is fed with synonymous sentences from different languages.
Precisely, we measure the pair-wise BLEU on the above IWSLT three-way-parallel test set. We choose one language as the target language, e.g., German, and then translate the other two languages, e.g., Italian and Dutch, to the target language. After obtaining these two translation files, we use one file as the reference, the other as the translation to calculate the BLEU, and then we swap the role of these two files to calculate the BLEU again. We average the BLEU scores to get the pair-wise BLEU, and the results in Table~\ref{tab:pbleu} show that our method can achieve higher results, which proves that our method can improve the prediction consistency.

\begin{table}[t]
\centering
\resizebox{\columnwidth}!{
\begin{tabular}{c|cccc}
\hline
System & IWSLT & IWSLT-b & PC-6 & OPUS-7 \\ \hline \hline
ZS & 93.2\% & 93.72\% & 87.93\% & 74.1\% \\
SRA & 93.9\% & 93.88\% & 91.54\% & 85.83\% \\
CL & 93.97\% & 93.96\% & 91.76\% & 86.23\% \\
Ours & \textbf{94.03\%} & \textbf{94.06\%} & \textbf{93.24\%} & \textbf{86.75\%} \\ \hline
\end{tabular}
}
\caption{The target language prediction accuracy.}
\label{tab:pa}
\end{table}

\subsection{Inspecting Spurious Correlations} 
The zero-shot translation usually suffers from capturing spurious correlations in the supervised directions, which means that the model overfits the mapping relationship from the input language to the output language observed in the training set~\cite{GuWCL19}.
%One of the reasons for the poor quality of zero-shot translation is that the model captures the spurious correlations in the dataset, which means that the model overfits the mapping relationship from the input language to the output language seen in the training set~\cite{GuWCL19}. 
This problem often causes the off-target prediction phenomenon where the model generates translation in the wrong target languages. To check whether our method can alleviate this phenomenon, we use the $\mathrm{Langdetect}$~\footnote{https://github.com/Mimino666/langdetect} toolkit to identify the target language and calculate the prediction accuracy as $1-n_{off-target}/n_{total}$.
We also compare our method with the 'SRA' and 'CL' methods. 
The results are shown in Table~\ref{tab:pa}. 
The 'ZS' baseline system can achieve high prediction accuracy on the IWSLT dataset, but the performance begin to decline as the amount of data becomes unbalanced and the languages become more unrelated. 
On all the datasets, our method achieves higher prediction accuracy and outperforms all the contrast methods. We can conclude from the results that our method can reduce the spurious correlation captured by the model.  

\section{Related Wrok}
%\textbf{Zero-shot Translation} 
Recent work on zero-shot translation can be divided into two categories. The first category helps the encoder produce language-agnostic features via extra regularization loss or training tasks. \citet{PhamNHW19} propose to compress the output of the encoder into a consistent number of states. \citet{abs-1903-07091} maximize the cosine similarities between the averaged representations of the source and target sentences. \citet{PanWWL20} and \citet{WeiW0XYL21} propose contrastive learning schemes to minimize the averaged sentence representation gap of similar sentences and maximize that of irrelevant sentences. Compared with their methods, we directly bridge the gap between two state sequences, which alleviates the mismatch problem of sentence representation.
\citet{JiZDZCL20} leverage explicit alignment information by external aligner tool or additional attention layer to obtain the aligned words for masking, and then they let the model predict the masked words based on the surrounding words. Compared with this work, our method is to align the whole state sequences of different languages, not just for single words. 
%Besides, our method does not introduce additional mask symbols which can avoids the problem of inconsistency brought by them.
\citet{LiuNCGL20} remove the residual connections in a middle layer of the encoder to release the positional correspondence to input tokens.
\citet{abs-2109-04705} introduce a denoising auto-encoder objective to improve the translation accuracy.
\citet{abs-2109-04778} leverage an auxiliary target language prediction task to retain information about the target languages. 
\citet{Yang22} uses optimal transport theory to improve the low-resource neural machine translation.
Compared with these work, our method introduces explicit constraints to the semantic representations.  

The second category extends the training data by generating pseudo sentence pairs or utilizing monolingual data. \citet{GuWCL19} apply decoder pre-training and back-translation to improve the zero-shot ability. \citet{Al-ShedivatP19} first translate the source and target languages to a third language and then make consistent predictions based on this pseudo sentence. \citet{ZhangWTS20} propose random online back translation to enforce the translation of unseen training language pairs. \citet{abs-2104-08757} fuse the pretrained multilingual model to the NMT model. Compared with these works, our method does not need additional data or additional time to generate pseudo corpus. If necessary, our method can also be combined with these works to further improve the zero-shot performance of the model.
\citet{YangYMHZLW20} propose to substitute some fragments of the source language with their counterpart translations to get the code-switch sentences. Compared to this work, our agreement-based method mixups the translation pairs to generate the pseudo sentence as the decoder input and then help the model to make consistent predictions. 

%Overall, our method belongs to the first category. The work of~\citet{abs-1903-07091,PanWWL20} are most related to us, which all try to align the cross-lingual semantic space further. Compared with their methods, we directly bridge the gap between two state sequences, which alleviates the mismatch problem of sentence representation. 

\section{Conclusion}
In this work, we focus on improving the zero-shot ability of multilingual neural machine translation. To reduce the discrepancy of the encoder output, we propose the state mover's distance based on the optimal transport theory and directly minimize the distance during training. We also propose an agreement-based training method to help the decoder make consistent predictions based on the semantic-equivalent sentences. The experimental results show that our method can get consistent improvements on diverse multilingual datasets. Further analysis shows that our method can
better align the semantic space, improve the prediction consistency, and reduce the spurious correlations. %captured by the model and help better align the semantic space.
%The many-to-many multilingual neural machine translation can perform zero-shot translation, which usually suffers poor translation quality. 
%To improve the zero-shot translation ability, some previous work propose to bridge the gap between the averaged sentence representations, which ignores some token-level information and also bring the mismatch problem. 
%we propose the state mover's distance based on the optimal transport theory and directly reduce the difference of the state sequences.    

\section*{Limitations}
Although our method can improve the performance of the zero-shot translation directions, it has limited benefits for the supervised translation performance.
%the improvements in supervised translation directions are limited.
%it does not bring significant improvements in the supervised translation directions. 
%We think that helping the model learn more language shared knowledge, e.g., the universal representations and cross-mappings, is limited to improving the supervised translation performance, 
On the one hand, the vanilla MNMT model has already been able to learn a lot of language shared knowledge. On the other hand, the language-specific knowledge learned by the model can also help the model achieve good translation performance in the supervised translation directions. 
Therefore, our method is limited to improving the supervised translation performance.
Besides, some reviewers pointed out that our method degraded the supervised translation performance according to the results of the main experiments.
This is because we select the checkpoints based on the performance of the zero-shot valid sets, which may cause a slight decline in the performance of the supervised directions. If we select checkpoints based on the the supervised valid sets, our method can improve the zero-shot performance without degrading the BLEU of the supervised directions. 
%For example, on the IWSLT dataset, if we select checkpoints based on the supervised valid sets, the average supervised BLEU of baseline is 30.81, and the zero-shot BLEU is 16.91. Meanwhile, the supervised BLEU of our method is 30.85, and the zero-shot BLEU is 18.55. 

\section*{Acknowledgements}
We thank all the anonymous reviewers for their insightful and valuable comments.

\bibliography{anthology,custom}
\bibliographystyle{acl_natbib}

\newpage
\appendix

\section{Appendix}
\label{sec:appendix}

% \usepackage{multirow}
\begin{table*}[t]
\centering

\subtable{
\begin{tabular}{c|cccccccccccc}
\hline
\multirow{2}{*}{PC-6} & Cs-Kk &  & Kk-Ru &  & Ro-Ru &  & Tr-Ro &  & Cs-Ro &  & Cs-Ru &  \\
 & $\rightarrow$ & $\leftarrow$ & $\rightarrow$ & $\leftarrow$ & $\rightarrow$ & $\leftarrow$ & $\rightarrow$ & $\leftarrow$ & $\rightarrow$ & $\leftarrow$ & $\rightarrow$ & $\leftarrow$ \\ \hline
PivT & 1.77 & 2.55 & 11.37 & 10.51 & 32.86 & 28.1 & 20.03 & 14.47 & 25.47 & 25.7 & 27.05 & 24.26 \\
ZS & 2.07 & 2.69 & 15.61 & 15.7 & 20.65 & 20.6 & 13.44 & 11.69 & 19.82 & 18.81 & 20.19 & 20.26 \\
SRA & 2.15 & 2.5 & 17.03 & 16.86 & 25.37 & 25.66 & 17.35 & 14.6 & 23.62 & 23.91 & 22.68 & 21.6 \\
CL & 1.99 & 2.68 & 16.48 & 16.49 & 29.28 & 26.8 & 17.82 & 15.66 & 23.87 & 23.42 & 27.05 & 24.29 \\
DisPos & \textbf{2.24} & 2.74 & 17.14 & \textbf{17.95} & 21.87 & 23.47 & 14.73 & 13.52 & 20.42 & 19.96 & 27.18 & 25.7 \\
DT & 2.2 & 2.87 & 19.23 & 18.88 & 28.05 & 25.88 & 17.82 & 14.41 & 22.29 & 22.3 & 26.29 & 23.98 \\
TLP & 2.01 & 2.82 & 14.59 & 13.01 & 28.41 & 25.88 & 18.53 & 13.25 & 23.11 & 22.54 & 25.24 & 22.74 \\
ZS+OT & 2.16 & 3.02 & 18.12 & 16.35 & \textbf{30.71} & \textbf{27.84} & 19.18 & 15.63 & \textbf{24.44} & 24.17 & 27.18 & \textbf{25.71} \\
ZS+AT & 2.06 & 2.82 & 15.8 & 16.54 & 28.01 & 26.37 & 19.25 & 15.59 & 22.63 & 22.55 & 24.6 & 23.27 \\
Ours & 2.2 & \textbf{3.08} & \textbf{18.3} & 17.91 & 30.59 & 27.73 & \textbf{19.66} & \textbf{16.16} & 23.58 & \textbf{24.49} & \textbf{27.22} & 25.66 \\ \hline
\end{tabular}
}

\subtable{
\begin{tabular}{c|ccccccccc}
\hline
\multirow{2}{*}{PC-6} & Cs-Tr &  & Kk-Ro &  & Kk-Tr &  & Ru-Tr &  & Zero \\
 & $\rightarrow$ & $\leftarrow$ & $\rightarrow$ & $\leftarrow$ & $\rightarrow$ & $\leftarrow$ & $\rightarrow$ & $\leftarrow$ & Avg. \\ \hline
PivT & 13.37 & 16.36 & 3.3 & 2.75 & 2.91 & 2.11 & 11.59 & 15.31 & 14.58 \\
ZS & 11 & 12.44 & 3.06 & 3.26 & 3.81 & 2.44 & 10.66 & 10.88 & 11.95 \\
SRA & 12.32 & \textbf{15.37} & 2.82 & 2.72 & 2.7 & 1.87 & 10.72 & 12.17 & 13.68 \\
CL & 12.02 & 14.16 & 3.33 & \textbf{3.34} & 3.49 & 2.44 & 11.72 & 13.52 & 14.49 \\
DisPos & 12.8 & 15.26 & 3.25 & 3.17 & 3.9 & 3.09 & 10.22 & 8.56 & 12.51 \\
DT & 11.62 & 13.38 & 3.49 & 3.37 & 3.96 & 3.24 & 11.97 & 13.38 & 14.43 \\
TLP & 11.96 & 13.98 & 3.33 & 2.98 & 3.65 & 2.98 & 12.02 & 13.5 & 13.83 \\
ZS+OT & 12.83 & 14.54 & \textbf{3.5} & 3.11 & 3.94 & \textbf{3.26} & 11.92 & 14.44 & 15.1 \\
ZS+AT & 11.99 & 14.11 & 3.41 & 3.03 & 3.46 & 2.54 & 11.9 & 14.11 & 14.2 \\
Ours & \textbf{12.85} & 15.21 & 3.24 & 3.18 & \textbf{3.95} & 3.04 & \textbf{12.81} & \textbf{14.96} & 15.29 \\ \hline
\end{tabular}
}

\caption{The results of each zero-shot translation direction on the PC-6 corpus. The notations denote the same meaning as in Table~\ref{tab:res}.}
\label{tab:pc6res}
\end{table*}

\subsection{PC-6 Data}

\begin{table}[t]
\centering
\begin{tabular}{l|l}
\hline
OPUS-6 & Size \\ \hline
En-Kk & 0.12M \\
En-Tr & 0.39M \\
En-Ro & 0.77M \\
En-Cs & 0.82M \\
En-Ru & 1.84M \\ \hline
\end{tabular}
\caption{The statistics about the PC-6 corpus.}
\label{tab:pc6}
\end{table}

The detailed statistics about the PC-6 corpus are shown in Table~\ref{tab:pc6}

\subsection{Experiments Results on PC-6}

The detailed results on the PC-6 corpus are shown in Table~\ref{tab:pc6res}.

\begin{table}[t]
\centering
\begin{tabular}{c|cc}
\hline
 & $\gamma_1$ & $\gamma_2$ \\ \hline
IWSLT & 0.4 & 0.001 \\
IWSLT-b & 0.2 & 0.002 \\
PC-6 & 0.2 & 0.003 \\
OPUS-7 & 0.3 & 0.01 \\ \hline
\end{tabular}
\caption{The hyperparameters $\gamma_1$ and $\gamma_2$ on each dataset. }
\label{tab:hyper}
\end{table}

\iffalse
\begin{figure*}[t]
    \centering
    \subfigure[ZS+OT system with different $\gamma_1$]{
        \includegraphics[width=\columnwidth]{gamma1.png}
    }
    \subfigure[ZS+AT system with different $\gamma_2$]{
        \includegraphics[width=\columnwidth]{gamma2.png}
    }
    \caption{ The BLEU of our method with different $\gamma_1$ and $\gamma_2$ on the IWSLT dataset. The BLEU of ZS baseline system is $17.01$.}
    \label{fig:hyper}
\end{figure*}
\fi

\begin{table}[t]
\centering
\begin{tabular}{cc|c}
\hline
$\alpha$ & $\beta$ & zero Avg. \\ \hline
1 & 1 & 17.23 \\
6 & 2 & 17.44 \\
6 & 3 & 17.59 \\
6 & 4 & 17.5 \\ \hline
\end{tabular}
\caption{The averaged BLEU with different $\alpha$ and $\beta$ for the 'ZS+AT' system. }
\label{tab:ab}
\end{table}

\begin{table}[t]
\centering
\begin{tabular}{c|cc}
\hline
 & kwps & ratio \\ \hline
ZS & 199 & 1 \\
SRA & 118 & 0.59 \\
SF & 61 & 0.31 \\
CL & 94 & 0.47 \\
ZS+OT & 125 & 0.63 \\
ZS+AT & 61 & 0.31 \\
Ours & 58 & 0.29 \\ \hline
\end{tabular}
\caption{The training speed on the IWSLT dataset.}
\label{tab:speed}
\end{table}

%\subsection{Implementation Details}
%All the systems are trained on 4 RTX3090 GPUs with the update frequency 2. The max token is 4096 for each GPU.
%We employ the Adam optimizer with $\beta_1 = 0.9$ and $\beta_2 = 0.98$. 
%We use the inverse square root learning scheduler and set the $warmup\_steps=4000$ and $lr=0.0007$. We set dropout s 0.1 for the PC-6 and OPUS-7 datasets. Since the IWSLT17 dataset is smaller than the other two datasets, we set the dropout rate as 0.3 for better performance. For the IWSLT data sets, we first pretrain the model for 20K steps and then continual-train the model with the proposed loss terms for 80K steps. For the PC-6 and OPUS-7 datasets, we first pretrain the model for 100K steps and then continual-train the model with the proposed loss terms for 200K steps. 

\subsection{Hyperparameters}

\noindent \textbf{$\gamma_1$ and $\gamma_2$} The hyperparameter $\gamma_1$ and $\gamma_2$ in Equation~\ref{eq:allloss} are set as in Table~\ref{tab:hyper}. 
%Besides, we conduct experiments on the IWSLT dataset with different hyperparameters. The results in Figure~\ref{fig:hyper} show that our method can always bring stable improvements. 

\noindent \textbf{$\alpha$ and $\beta$} We tried several combinations of $\alpha$ and $\beta$, and report the averaged BLEU in Table. Under the optimal setting ($\alpha=6, \beta=3$), the probability expectation that the words of the pseudo sentence $\mathbf{z}$ come from the source sentence $\mathbf{x}$ is $0.67$ and from the target sentence $\mathbf{y}$ is $0.33$.

%\subsection{Ablation Study} For the 'ZS+OT' system, we add use 

\subsection{Training Speed} 
We test the training speed of all the systems. All the speeds are measured as kilo-words per second (kwps) and tested in parallel on 4 RTX3090 GPUs with the same max token and update frequency. We also report the speed ratios of different systems compared with the speed of the ZS system. The results are shown in Table~\ref{tab:speed}. The results show that our 'ZS+OT' system is faster than the 'SRA' and 'CL' systems with better performance. The 'ZS+AT' system is much slower because it needs three complete forward propagations.

\end{document}